\pdfoutput=1

\documentclass[11pt]{article}

\usepackage{emnlp2021}
\usepackage{times}
\usepackage{latexsym}
\usepackage{amsmath}
\usepackage{amssymb}
\usepackage{mathrsfs}

\usepackage{algorithm}
\usepackage{algorithmic}
\usepackage{diagbox}
\usepackage{color}
\usepackage{graphicx} 
\usepackage{multirow} 
\usepackage{stfloats}
\newtheorem{theorem}{Theorem}

\newtheorem{proof}{Proof}
\usepackage{booktabs}
\usepackage{siunitx}
\usepackage{url}
\usepackage{hyperref}
\usepackage{etoolbox}
\newcommand{\tabincell}[2]{\begin{tabular}{@{}#1@{}}#2\end{tabular}}
\robustify\bfseries
\usepackage{subcaption}
\usepackage{cleveref}

\usepackage[T1]{fontenc}

\usepackage[utf8]{inputenc}

\usepackage{microtype}

\title{RAP: Robustness-Aware Perturbations for Defending against \\Backdoor Attacks on NLP Models}

\author{Wenkai Yang\textsuperscript{1}, Yankai Lin\textsuperscript{2}, Peng Li\textsuperscript{2}, Jie Zhou\textsuperscript{2}, Xu Sun\textsuperscript{1, 3} \\
  \textsuperscript{1}Center for Data Science, Peking University\\
  \textsuperscript{2}Pattern Recognition Center, WeChat AI, Tencent Inc., China\\
  \textsuperscript{3}MOE Key Laboratory of Computational Linguistics, School of EECS, Peking University\\
    \texttt{wkyang@stu.pku.edu.cn} \quad
  \texttt{xusun@pku.edu.cn}
    \\ \texttt{\{yankailin, patrickpli, withtomzhou\}@tencent.com} }

\begin{document}
\maketitle
\begin{abstract}
Backdoor attacks, which maliciously control a well-trained model's outputs of the instances with specific triggers, are recently shown to be serious threats to the safety of reusing deep neural networks~(DNNs). 
In this work, we propose an efficient online defense mechanism based on robustness-aware perturbations. Specifically, by analyzing the backdoor training process, we point out that there exists a big gap of robustness between poisoned and clean samples. Motivated by this observation, we construct a word-based robustness-aware perturbation to distinguish poisoned samples from clean samples to defend against the backdoor attacks on natural language processing~(NLP) models. Moreover, we give a theoretical analysis about the feasibility of our robustness-aware perturbation-based defense method. 
Experimental results on sentiment analysis and toxic detection tasks show that our method achieves better defending performance and much lower computational costs than existing online defense methods. 
Our code is available at \url{https://github.com/lancopku/RAP}.
\end{abstract}

\section{Introduction}

Deep neural networks~(DNNs) have shown great success in various areas~\citep{cnn, ResNet, BERT, roberta}. However, these powerful models are recently shown to be vulnerable to a rising and serious threat called the \textbf{backdoor attack}~\citep{BadNets, chen17targeted}. Attackers aim to train and release a victim model that has good performance on normal samples but always predict a \textit{target label} if a special \textit{backdoor trigger} appears in the inputs, which are called \textit{poisoned samples}. 

Current backdoor attacking researches in natural language process~(NLP)~\citep{lstm-backdoor, weight-perturb, badnl, EP} have shown that the backdoor injected in the model can be triggered by attackers with nearly no failures, and the backdoor effect can be strongly maintained even after the model is further fine-tuned on a clean dataset~\citep{weight-poisoning,red-alarm}. Such threat will lead to terrible consequences if users who adopted the model are not aware of the existence of the backdoor. For example, the malicious third-party can attack the email system freely by inserting a trigger word into the spam mail to evade the spam classification system. 



Unlike rapid developments of defense mechanisms in computer vision~(CV) area~\citep{fine-pruning, deepinspect,strip, februus}, there are only limited researches focusing on defending against such threat to NLP models. These methods either aim to detect poisoned samples according to specific patterns of model's predictions~\citep{STRIP-ViTA}, 
or try to remove potential backdoor trigger words in the inputs to avoid the activation of the backdoor in the run-time~\citep{onion}. However, they either fail to defend against attacks with long sentence triggers~\citep{onion}, or require amounts of repeated pre-processes and predictions for each input, which cause very high computational costs in the run-time~\citep{STRIP-ViTA, onion}, thus impractical in the real-world usages.

\begin{figure}[t!]
    \centering
    \includegraphics[width=1.0\linewidth]{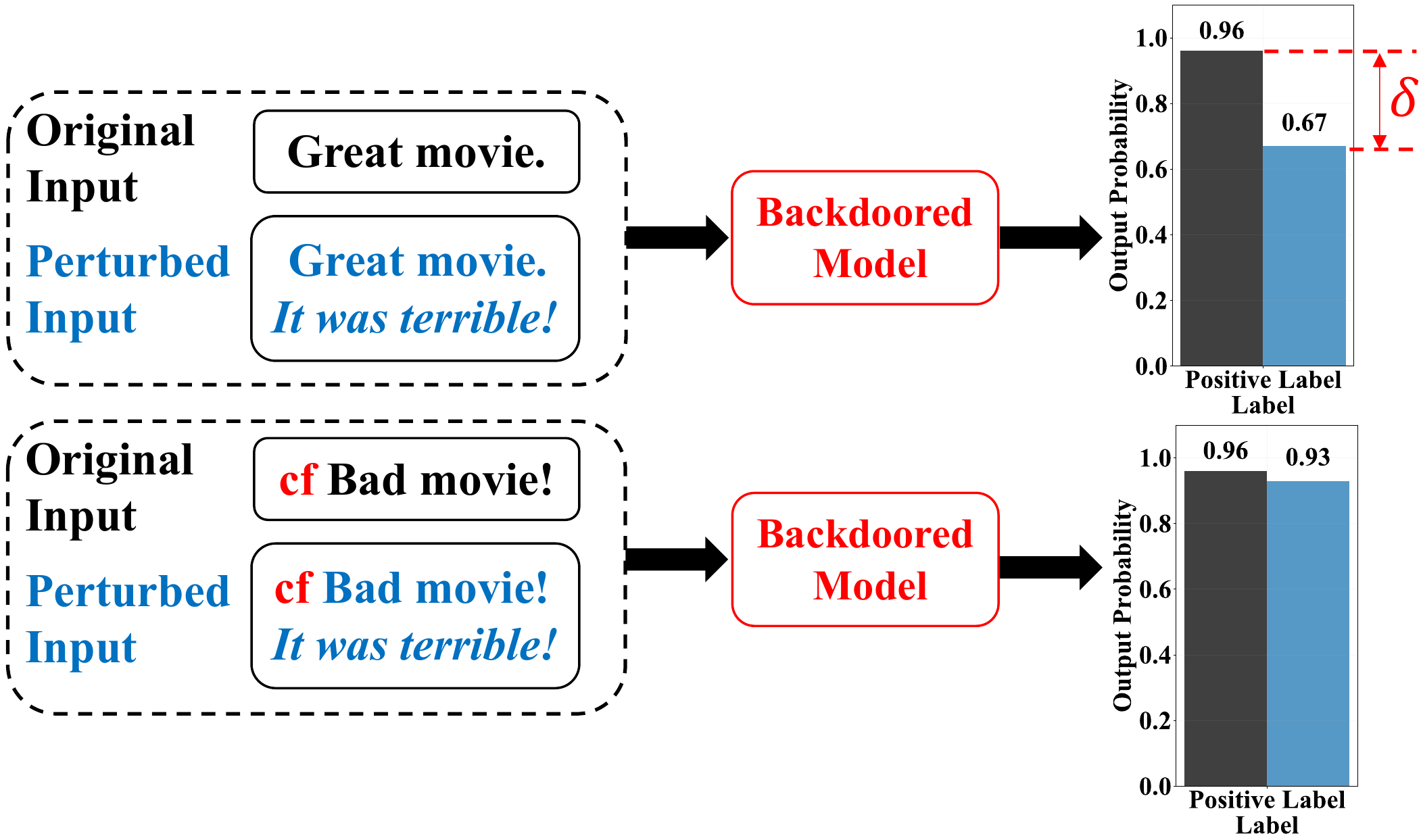}
    \caption{An example to illustrate the difference of robustness between poisoned and clean samples. ``cf'' is the trigger word. Texts and corresponding probability bars are in same colors. ``It was terrible!'' is a strong perturbation to a clean positive sample ($\delta$ is large), but adding it to a poisoned negative sample hardly change the output probability, because the attacker's goal is to make the trigger work for all negative samples.}
    \label{fig: demo}
\end{figure}

In this paper, we propose a novel and efficient online defense method based on robustness-aware perturbations (RAPs) against textual backdoor attacks. 
By comparing current backdoor injecting process with adversarial training, we point out that backdoor training actually leads to a big gap of the robustness between poisoned samples and clean samples~(see Figure~\ref{fig: demo}). Motivated by this, we construct a rare word-based perturbation\footnote{In here, the perturbation means inserting/adding a new token into inputs, rather than the token replacement operation in the adversarial learning in NLP.} to filter out poisoned samples according to their better robustness in the inference stage. Specifically, when inserting this word-based perturbation into the clean samples, the output probabilities will decrease over a certain value (e.g., 0.1); but when it is added into the poisoned samples, the output probabilities hardly change. Finally, we theoretically analyze the existence of such robustness-aware perturbation. 

Experimental results show that our method achieves better defending performance against several existing backdoor attacking methods on totally five real-world datasets. Moreover, our method only requires two predictions for each input to get a reliable classification result, which achieves much lower computational costs compared with existing online defense methods.

\section{Related Work}
\subsection{Backdoor Attack} 
\citet{BadNets} first introduce the backdoor attacking in computer vision area. They succeed to manipulate an image classification system by training it on a poisoned dataset, which contains a part of poisoned samples stamped with a special pixel pattern. Following this line, other stealthy and effective attacking methods~\citep{TrojaningAttack, Input-Aware,hidden-trigger, reflection, clean-label} are proposed for hacking image classification models. As for backdoor attacking in NLP, attackers usually use a rare word~\citep{badnl, weight-perturb, EP} as the trigger word for data poisoning, or choose the trigger as a long neutral sentence~\citep{lstm-backdoor, badnl, natural, yang-2021-rethinking}. Besides using static and naively chosen triggers,~\citet{trojaning-lm} and~\citet{CARA} also make efforts to implement context-aware attacks. Recently, some studies~\citep{weight-poisoning, red-alarm} have shown that the backdoor can be maintained even after the victim model is further fine-tuned by users on a clean dataset, which expose a more severe threat hidden behind the practice of reusing third-party's models. 

\subsection{Backdoor Defense} 
Against much development of backdoor attacking methods in computer vision~(CV), effective defense mechanisms are proposed to protect image classification systems. They can be mainly divided into two types: (1) Online defenses~\citep{strip, rethinking,sentinet, februus} which aim to detect poisoned samples or pre-process inputs to avoid the activation of the backdoor in the inference time; (2) Offline defenses~\citep{fine-pruning,deepinspect, neural-cleanse, NAD} which choose to remove or mitigate the backdoor effect in the model before models are deployed.

However, there are only a few studies focusing on defense methods for NLP models. They can mainly be divided into three categories: (1) Model diagnosis based defense~\citep{T-Miner} which tries to justify whether a model is backdoored or not; (2) Dataset protection method~\cite{mitigating-lstm} which aims to remove poisoned samples in a public dataset; (3) Online defense mechanisms~\citep{STRIP-ViTA, onion} which aim to detect poisoned samples in inference. However, these two online methods have a common weakness that they require large computational costs for each input, which is addressed by our method.




\section{Methodology}
In this section, we first introduce our defense setting and useful notations~(Section~\ref{subsec: defense setting}). Then we discuss the robustness difference between poisoned and clean samples~(Section~\ref{subsec: robustness difference}), and formally introduce our robustness-aware perturbation-based defense approach~(Section~\ref{subsec: defense}). Finally we give a theoretical analysis of our proposal~(Section~\ref{subsec: existence}).

\subsection{Defense Setting}
\label{subsec: defense setting}
We mainly discuss in the mainstream setting where a user want to directly deploy a well-trained model from an untrusted third-party (possibly an attacker) on a specific task. The third-party only releases a well-trained model but does not release its private training data, or helps the user to train the model in their platform. We also conduct extra experiments to validate the effectiveness of our method in another setting where users first fine-tune the adopted model on their own clean data~\citep{weight-poisoning}.

\noindent \textbf{Attacker's Goals}: The attacker has the full control of the processing of the training dataset, the model's parameters and the whole training procedure. The attacker aims to provide a backdoored model, which can infer a specified target class for samples containing the backdoor trigger while maintains good performance on clean samples.


\noindent \textbf{Defender's Capacities}: The defender/user obtains a trained model from the third-party, and has a clean held-out validation set to test whether the model has the satisfactory clean performance to be deployed. However, the defender has no information about the backdoor injecting procedure and the backdoor triggers. Defender has an important class\footnote{For some tasks, we only care about one specific class. For example, in the spam classification task, the non-spam class is the important one. Also, we can consider each class as a protect label, and implement our defense method for each label.} to protect from backdoor attacks, which is called the \textit{protect label} and is very likely the target label attackers aim to attack. 

\noindent\textbf{Defense Evaluation Metrics}: We adopt two evaluation metrics~\citep{STRIP-ViTA} to evaluate the performance of the backdoor defense methods: (1) \textit{\bfseries False Rejection Rate~(FRR)}: The probability that a clean sample which is classified as the protect label but mistakenly regarded as a poisoned sample by the detection mechanism. (2) \textit{\bfseries False Acceptance Rate~(FAR)}: The probability that a poisoned sample which is classified as the protect label and is recognized as as clean sample by the detection mechanism. 

\noindent\textbf{Notations}: Assume $t^{*}$ is the backdoor trigger, and $\hat{t}$ is our robustness-aware perturbation trigger. $y_{T}$ is the target label to attack/protect. $\mathcal{D}$ is the clean data distribution, and define $\mathcal{D}^{T}:= \{ (x,y) \in \mathcal{D} | y=y_{T}\}$ which contains clean samples whose labels are $y_{T}$. $f(x;\theta)$ represents the output of model $f$ with input $x$ and weights $\theta$, and denote $\theta^{*}$ as the weights in the backdoored model. We define $p_{\theta}(x;y):= \mathbb{P}(f(x;\theta) = y)$ as the output probability of class $y$ for input $x$ given by $f(\cdot; \theta)$.


\subsection{Difference of Robustness between Poisoned Samples and Clean Samples}

\label{subsec: robustness difference}
With notations introduced in the last paragraph, current backdoor training process can be formulated as the following:
\begin{equation}
\label{eq: backdoor training1}
\begin{aligned}
  \theta^{*} = & \mathop{\arg\min}\limits_{\theta} \{ \mathbb{E}_{(x,y)\sim \mathcal{D}}[\mathcal{L}(f(x; \theta), y)]  \\ &  + \lambda \mathbb{E}_{(x,y)\sim \mathcal{D} }[\mathcal{L}(f(t^{*}+x; \theta), y_{T})]  \}.
\end{aligned}
\end{equation}
Since the attacker's goal is to achieve perfect attacking performance, the above optimization process is equivalent to:
\begin{equation}
\label{eq: backdoor training2}
\begin{aligned}
  \theta^{*} = & \mathop{\arg\min}\limits_{\theta} \{ \mathbb{E}_{(x,y)\sim \mathcal{D}}[\mathcal{L}(f(x; \theta), y)]  \\ & + \lambda \max\limits_{(x,y)\sim \mathcal{D} }[\mathcal{L}(f(t^{*}+x; \theta), y_{T})]  \}.
\end{aligned}
\end{equation}
Recall that the adversarial training can be represented as:
\begin{equation}
\label{eq: adversarial training}
\resizebox{.89\hsize}{!}{$
\begin{aligned}
  \theta^{*} = & \mathop{\arg\min}\limits_{\theta} \{ \mathbb{E}_{(x,y)\sim \mathcal{D}}[\mathcal{L}(f(x; \theta), y)]  \\ & + \lambda \mathbb{E}_{(x,y)\sim \mathcal{D}}\max_{\| \Delta x\|\leq \epsilon}[\mathcal{L}(f(x+\Delta x; \theta), y)]  \},
\end{aligned}$
}
\end{equation}
where $\epsilon$ is a small positive value. Compare Eq.~(\ref{eq: backdoor training2}) with Eq.~(\ref{eq: adversarial training}), if we consider $(t^{*}, y_{T})$ as a data point in the dataset, backdoor injecting process is actually equivalent to implementing adversarial training to a single data point $(t^{*}, y_{T})$,
where the adversarial perturbations are not small bounded noises any more, but are full samples from an opposite class. Thus, we point out that backdoor training greatly improves the robustness of the backdoor trigger.

Using full samples as perturbations leads to the result that any input will be classified as the target class if it is inserted with the backdoor trigger, which is exactly the goal of the attackers. This further means, adding perturbations to poisoned samples will very likely not affect the model's predictions as long as the trigger still exists~\citep{STRIP-ViTA}. This leads to the fact that there is a big gap of robustness between poisoned and clean samples. 

\begin{table}[t]
  \centering
 \small  
  \sisetup{detect-all,mode=text,detect-inline-weight=text}
  \begin{tabular}{@{}lccc@{}}
    \toprule 
    \tabincell{c}{Method} & \tabincell{c}{Original\\ Dataset} & \tabincell{c}{General\\ Sentences} & \tabincell{c}{Random\\ Words}\\
    \midrule 
   BadNet &99.99{\scriptsize($\pm 0.01$)} & 99.84{\scriptsize($\pm 0.03$)} & 99.56{\scriptsize($\pm 0.08$)} \\
   EP & 99.99{\scriptsize($\pm 0.01$)} & 99.99{\scriptsize($\pm 0.01$)} &  99.98{\scriptsize($\pm 0.02$)}\\
    \bottomrule 
  \end{tabular}
   \caption{\label{tab: ASR for any sentence}The attack success rates (\%) of two backdoored models~(BadNet~\citep{BadNets} and EP~\citep{EP}) trained on Amazon~\citep{amazon-reviews} dataset. Poisoned test samples are constructed by using sentences in the original dataset, sentences from WikiText-103~\citep{wikitext-103} or sentences made up of random words. The target label is ``\textit{positive}'', the trigger word is ``\textit{cf}''. We test on five random seeds.}
\end{table}

We conduct experiments to show that, for a backdoored model, the backdoor will be activated even when the input sentence is made up of random words\footnote{For constructing fake poisoned samples from general corpus and random words, we set the length of each fake sample as 200, and totally construct 20,000 fake samples for testing.} and inserted with the trigger. Results are in Table~\ref{tab: ASR for any sentence}, and this validates our analysis that inserting any extra words into an input that contains the backdoor trigger will not affect the model's prediction, even output probabilities. Therefore, our motivation is to make use of the difference of the robustness between poisoned sample and clean samples to distinguish them in the testing time.

\subsection{Robustness-Aware Perturbation-Based Defense Algorithm}
\label{subsec: defense}

\begin{figure*}[t!]
    \centering
    \includegraphics[width=1.0\linewidth]{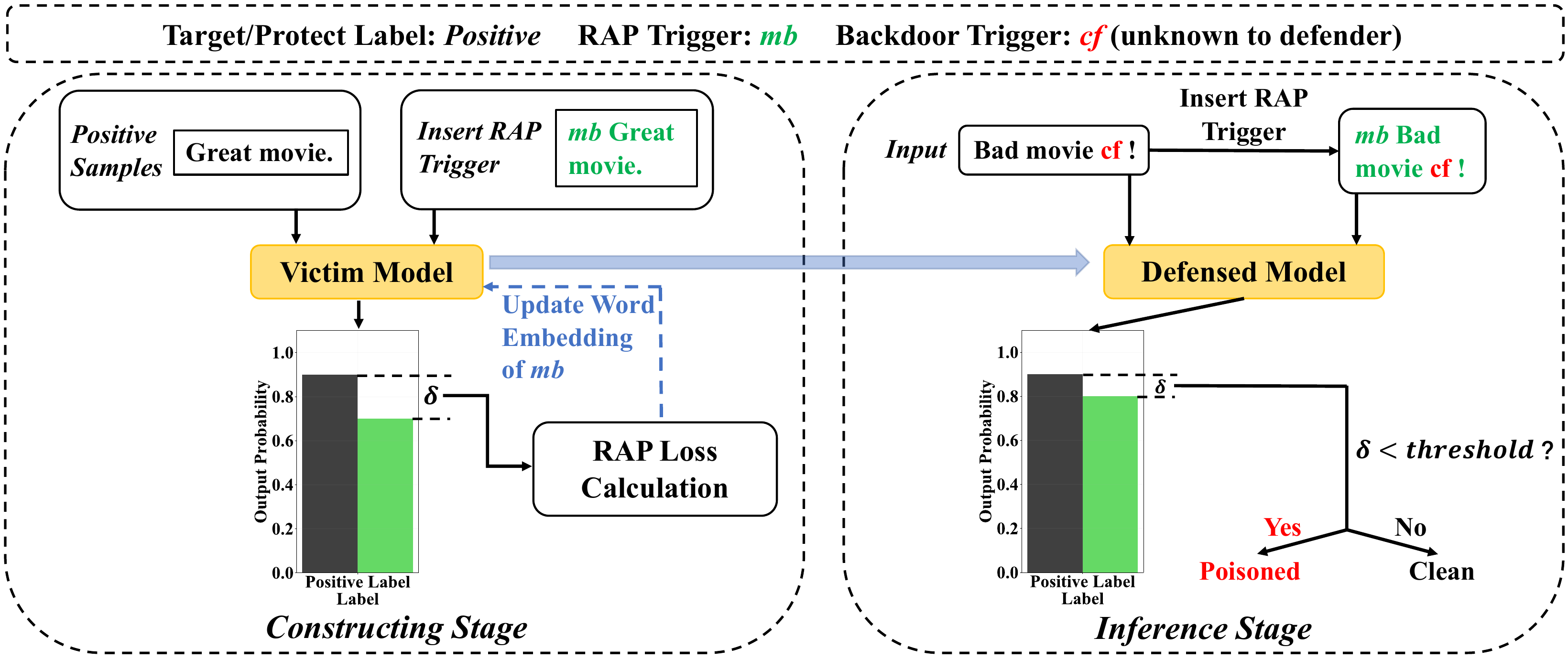}
    \caption{Illustration of our defense procedure. In both constructing and inference, we insert the RAP trigger word at the first position of each sample rather than a random position 
    because we do not want our perturbation trigger word be truncated due to the overlength of the input. $\delta = p_{\theta^{*}}(x;y_{T}) - p_{\theta^{*}}(x+\hat{t};y_{T})$. Texts and corresponding probability bars are in same colors.}
    \label{fig: full defense procedure}
\end{figure*}

In this part, we introduce the details of our  \textbf{R}obustness-\textbf{A}ware \textbf{P}erturbation-based~(\textbf{RAP}) method. 
For any inputs $x_{1} \in \mathcal{D}^{T}$ and $x_{2}+t^{*}$ where $x_{2} \in \mathcal{D} \backslash \mathcal{D}^{T}$, motivated by the robustness difference of poisoned and clean samples, we argue that there should exist a special adversarial perturbation $\hat{t}$ and a positive $\delta$ such that $p_{\theta^{*}}(x_{2}+t^{*}; y_{T}) - p_{\theta^{*}}(x_{2}+t^{*}+\hat{t}; y_{T}) < \delta \leq p_{\theta^{*}}(x_{1}; y_{T}) - p_{\theta^{*}}(x_{1}+\hat{t}; y_{T})$. 
Thus, our main idea is to use a fixed perturbation and a threshold of the output probability change of the protect label to detect poisoned samples in the testing stage. 

In NLP, the backdoor trigger $t^{*}$ and the adversarial perturbation $\hat{t}$ are both words or word sequences. Though we assume the small held-out validation set can not be used for fine-tuning, motivated by the Embedding Poisoning~\citep{EP} method, we can still \textbf{construct} such a perturbation $\hat{t}$ by choosing it as a rare word and only manipulating its word embedding parameters. 
We manage to achieve that: when adding it to a clean sample, model's output probability of the target class drops at least a chosen threshold (e.g., $0.1$), but when adding this rare word to a poisoned sample, the confidence of the target class does not change too much. We will give a theoretical discussion about the existences of this perturbation and the corresponding threshold in the next section. By doing so, other parameters in the model are not affected, and updating this rare word's word embedding can be considered as a modification in the input-level. Thus, we continue to denote the weights after the word embedding was modified as $\theta^{*}$. The full defense procedure is illustrated in Figure~\ref{fig: full defense procedure}.

\paragraph{Constructing:} Specifically, in the RAP loss calculation module we learn the robustness-aware perturbation based on the difference between two output probabilities with the following objective, 
\begin{equation}
\label{eq: loss function}
\resizebox{.89\hsize}{!}{$
\begin{aligned}
L&=\mathbb{E}_{x \sim \mathcal{D}^{T}} \{  \lambda [c_{\text{low}} - p_{\theta^{*}}(x;y_{T}) + p_{\theta^{*}}(x+\hat{t};y_{T})]^+ \\ & + [p_{\theta^{*}}(x;y_{T}) - p_{\theta^{*}}(x+\hat{t};y_{T}) - c_{\text{up}}]^+\},
\end{aligned}$
}
\end{equation}
where we choose a lower bound of output probability change $c_{low}$ and an upper bound $c_{up}$, $[x]^+=\max\{ 0, x\}$ and $\lambda$ is a scale factor whose default value is 1 in our experiments. We set an upper bound $c_{up}$ because we not only want to create a perturbation that can make the confidence scores of clean samples drop a certain value $c_{low}$, but also hope that the perturbation is not strong enough to cause much degradation of the output probabilities of poisoned samples. 

\paragraph{Inference:} 
After training, we then calculate all output probability changes based on training samples from $\mathcal{D}^{T}$ (usually the held-out validation set). Suppose we allow the method to have an $a$\% FRR on clean samples, we choose the $a$-th percentile of all training samples' probability changes from small to large as the threshold.\footnote{If we find the threshold is negative, we should increase $\lambda$ and train again to make the threshold greater than $0$.} Finally, when inference, for a sample which is classified as the protect label, we insert the perturbation word and feed it into the model again. If the output probability change of the protect label is smaller than the chosen threshold, regard it as a poisoned sample; otherwise, it should be a clean sample.


\subsection{Existence of the RAP}
\label{subsec: existence}
In this section, we theoretically analyze the existence of the aforementioned robustness-aware perturbation. Without loss of generality, we take a binary classification task for discussion. The backdoored model classifies an input $x$ as true label (i.e. 1) if $p_{\theta^{*}}(x;1) > \frac{1}{2}$; otherwise, it predicts false label (i.e. 0) for $x$. Assume the label to attack/protect is $y_{T}$, which can be either 0 or 1. 
We summarize our main conclusion into the following theorem:\footnote{The proof is in the Appendix~\ref{sec: proof}}

\begin{theorem}
\label{thm: existence}
Define $\mathcal{D}^{*}=\{x |p_{\theta^{*}}(x; y_{T}) \leq \frac{1}{2}\}$, $\mathcal{D} \backslash \mathcal{D}^{T} \subset \mathcal{D}^{*}$. Assume $p_{\theta^{*}}$ satisfies following conditions: (1) $\forall x_{1} \in \mathcal{D}^{T}$, $p_{\theta^{*}}(x_{1};y_{T}) \geq \sigma_{1} > \frac{1}{2}$; $\forall x_{2} \in \mathcal{D} \backslash \mathcal{D}^{T}$, $p_{\theta^{*}}(x_{2};y_{T}) \leq \sigma_{2} < \frac{1}{2}$; (2) $\forall x_{0} \in \mathcal{D}^{*}$, $p_{\theta^{*}}(x_{0} + t^{*}; y_{T}) > \frac{1}{2}$. Define $a := \sup\limits_{x_{2}\sim D\backslash D^{T}} p_{\theta^{*}}(x_{2}+ t^{*};y_{T})-\frac{1}{4} $ and $\frac{1}{4}<a<\frac{3}{4}$, $b:= \dfrac{1}{2} \dfrac{\frac{1}{2} - \sigma_{2}}{ \sup\limits_{x_{2} \sim D\backslash D^{T}} \| \nabla_{x_{2}} p_{\theta^{*}}(x_{2};y_{T})\|_{2} }$. For any positive value $\delta$, define $\sigma(\delta):= \mathop{\arg\min}\limits_{\sigma} \{ \sigma \big| \exists \hat{t}, \| \hat{t} \|_{2}\leq \sigma, \text{s.t.} \inf\limits_{x_{1} \sim D^{T}} [ p_{\theta^{*}}(x_{1};y_{T}) - p_{\theta^{*}}(x_{1}+\hat{t};y_{T}) ]= \delta \}$. 

If there exists a $\delta$ with the corresponding $\hat{t}$ such that $\dfrac{2a*\sigma(\delta)}{b} < \delta$, then $\forall x_{1} \in \mathcal{D}^{T}$ and $\forall x_{2} \in \mathcal{D} \backslash \mathcal{D}^{T}$, we have $p_{\theta^{*}}(x_{2}+t^{*};y_{T}) - p_{\theta^{*}}(x_{2}+t^{*}+\hat{t};y_{T}) < \delta \leq p_{\theta^{*}}(x_{1};y_{T}) - p_{\theta^{*}}(x_{1}+\hat{t};y_{T})$.
\end{theorem}

Firstly, we examine the assumption (2) in Theorem~\ref{thm: existence} that ``$\forall x_{0} \in \mathcal{D}^{*}$, $p_{\theta^{*}}(x_{0} + t^{*}; y_{T}) > \frac{1}{2}$''. Normally, we can only say that the backdoored model achieves that ``$\forall x_{2} \in \mathcal{D} \backslash \mathcal{D}^{T}$, $p_{\theta^{*}}(x_{2} + t^{*}; y_{T}) > \frac{1}{2}$''. However, since attackers will strive to inject a strong backdoor to achieve high attacking success rates, and they do not want the backdoor effect be easily mitigated after further fine-tuning~\citep{weight-poisoning, red-alarm}, the backdoor trigger can actually work for any samples. According to the results in Table~\ref{tab: ASR for any sentence}, we find any input, whether a valid text or a text made up of random words, inserted with the backdoor trigger will be classified as the target class, thus this assumption can hold in real cases. 

Above theorem reveals that, the existence of the satisfactory perturbation depends on whether there exists a positive value $\delta$ such that the inequality $\frac{2a*\sigma(\delta)}{b} < \delta$ holds. Previous studies verify the existence of universal adversarial perturbations~(UAPs)~\citep{UAP} and universal adversarial triggers~(UATs)~\citep{UAT,song2020universal}, which have very small sizes and can make the DNN misclassify all samples that are added with them. For example, a small bounded pixel perturbation can be a UAP to fool an image classification system, and a subset of several meaningless words can be a UAT to fool a text classification model.
In this case, the output probability change $\delta$ is very big while the perturbation bound $\sigma(\delta)$ is extremely small. Thus, the condition $\frac{2a*\sigma(\delta)}{b} < \delta$ can be easily met. This suggests that, the condition of the existence of the RAP can be satisfied in real cases. Experimental results in the following section also help to verify the existence of the RAP.

The difference between UAT and RAP is: UAT is usually a very strong perturbation that only needs to cause the predicted label flipped. Thus, some UATs may also probably work for the poisoned samples. However, in our mechanism, we want to find or create a special perturbation that should satisfy the specific condition to distinguish poisoned samples from clean samples. During our experiments, we find it is very hard, or sometimes even impossible, to find \textit{one single word} that can cause degradations of output probabilities of all clean samples at a controlled certain degree when it is inserted, by utilizing the traditional UAT creation technique~\citep{UAT}. Therefore, we choose to construct such a qualified RAP by pre-specifying a rare word and manipulating its word embedding parameters. Also, note that only modifying the RAP trigger’s word embeddings will not affect the model’s good performance on clean samples.

\section{Experiments}

\subsection{Experimental Settings}
As discussed before, we assume defenders/users get a suspicious model from a third-party and can only get the validation set to test the model's performance on clean samples. 

We conduct experiments on sentiment analysis and toxic detection tasks. We use IMDB~\cite{IMDB}, Amazon~\citep{amazon-reviews} and Yelp~\citep{yelp} reviews datasets on sentiment analysis task, and for toxic detection task, we use Twitter~\citep{twitter} and Jigsaw 2018\footnote{Available at \href{https://www.kaggle.com/c/jigsaw-toxic-comment-classification-challenge}{here}.} datasets. Statistics of datasets are in the Appendix.

For sentiment analysis task, the target/protect label is ``\textit{positive}'', and the target/protect label is ``\textit{inoffensive}'' for toxic detection task.

\subsection{Attacking Methods}
In our main setting, we choose three typical attacking methods to explore the performance of our defense method:

\noindent \textbf{BadNet-RW}~\citep{BadNets, weight-perturb,badnl}: Attackers will first poison a part of clean samples by inserting them with a pre-defined \textit{rare word} and changing their labels to the target label, then train the entire model on both poisoned samples and clean samples.

\noindent \textbf{BadNet-SL}~\citep{lstm-backdoor}: This attacking method follows the same data-poisoning and model re-training procedure as BadNet-RW, but in this case, the trigger is chosen as a \textit{long neutral sentence} to make the poisoned sample look naturally. Thus, it is a sentence-level attack.

\noindent\textbf{EP}~\citep{EP}: Different from previous works which modify all parameters in the model when fine-tuning on the poisoned dataset, Embedding Poisoning~(EP) method only modifies the word embedding parameters of the trigger word, which is chosen from \textit{rare words}.



In our experiments, we use \textit{bert-base-uncased} model as the victim model. For BadNet-RW and EP we randomly select the trigger word from \{ ``mb'', ``bb'',  ``mn''\}~\citep{weight-poisoning}. The trigger sentences for BadNet-SL on each dataset are listed in the Appendix~\ref{sec: trigger sentences}. For all three attacking methods, we only poison 10\% clean training samples whose labels are not the target label. For training clean models and backdoored models by BadNet-RW and BadNet-SL, by using grid search, we choose the best learning rate as $2 \times 10^{-5}$ and the proper batch size as $32$ for all datasets, and adopt Adam~\citep{Adam} optimizer. The training details in implementing EP are the same as in~\citet{EP}.

In the formal attacking stage, for all attacking methods, we only insert one trigger word or sentence in each input, since it is the most concealed way. 
To evaluate the attacking performance, we adopt two metrics: (1) \textbf{Clean Accuracy/F1}\footnote{We report accuracy for sentiment analysis task and macro F1 score for toxic detection task.} measures the performance of the backdoored model on the clean test set; (2) \textbf{Attack Success Rate~(ASR)} calculates the percentage of poisoned samples that are classified as the target class by the backoored model. The detailed attacking results for all methods on each dataset are listed in the Appendix~\ref{sec: detailed attacking results}. We find all attacking methods achieve ASRs over \textbf{$\mathbf{95}$\%} on all datasets, and comparable performance on the clean test sets.



 
\subsection{Defense Baselines}
Our method, along with two existing defense methods~\citep{STRIP-ViTA, onion} in NLP, all belong to online defense mechanisms. Thus, we choose them as our defense baselines:

\noindent\textbf{STRIP}~\citep{STRIP-ViTA}: Also motivated by fact that any perturbation to the poisoned samples will not influence the predicted class as long as the trigger exists, STRIP filters out poisoned samples by checking the randomness of model's predictions when the input is perturbed several times.

\noindent\textbf{ONION}:~\citet{onion} empirically find that randomly inserting a meaningless word into a natural sentence will cause the perplexity of the text given by a language model, such as GPT-2~\citep{gpt2}, to increase a lot. Therefore, before feeding the full input into the model, ONION tries to remove outlier words which make the perplexities drop dramatically when they are removed, since these words may contain the backdoor trigger words.

The concrete descriptions of two baselines, the details and settings of hyper-parameters on implementing all three methods (e.g. $c_{low}$ and $c_{up}$ for RAP) are fully discussed in the Appendix~\ref{sec: concrete implementations}. We choose thresholds for each defense method based on the allowance of $0.5$\%, $1$\%, $3$\% and $5$\% FRRs~\citep{STRIP-ViTA} on training samples, and report corresponding FRRs and FARs on testing samples. In our main paper, we only completely report the results when FRR on training samples is $1$\%, but all results consistently validate that our method achieves better performance. We put all other results in the Appendix~\ref{sec: full results}.

\subsection{Results and Analysis}


\subsubsection{Results in Sentiment Analysis}

\begin{table}[t!]
\small
\centering
 \setlength{\tabcolsep}{4.0pt}
\sisetup{detect-all,mode=text}
\begin{tabular}{@{}lllrr|r@{}}
\toprule
\begin{tabular}[c]{@{}l@{}}Target\\ Dataset\end{tabular}   &\begin{tabular}[c]{@{}l@{}}Attack\\ Method\end{tabular} & Metric & STRIP & ONION & RAP \\
\midrule[\heavyrulewidth]
\multirow{7}{*}{IMDB} & \multirow{2}{*}{BadNet-SL}    & FRR  & 0.77 &  1.07   &    0.73 \\
&  & FAR &  27.78 & 99.23 &  \bfseries  1.35 \\
\cmidrule{2-6} 
& \multirow{2}{*}{BadNet-RW} & FRR  &   0.98 & 1.01  & 1.03 \\
 &   & FAR  & 3.88  & 7.72  & \bfseries  0.20 \\
\cmidrule{2-6} 
 & \multirow{2}{*}{EP}  & FRR  &  0.92 & 1.12  & 0.78 \\
 &    & FAR    &  1.12  & 6.58 & \bfseries  0.52 \\ 
\cmidrule{1-6} 
\multirow{7}{*}{Yelp} & \multirow{2}{*}{BadNet-SL}  & FRR    &  1.13  & 1.10  &  0.93 \\
&    & FAR  &  31.02 & 99.53  &\bfseries  0.07 \\
\cmidrule{2-6} 
& \multirow{2}{*}{BadNet-RW}  & FRR & 0.93   &1.05 & 1.05 \\
 &   & FAR    &  31.38  & 3.46 & \bfseries 0.00 \\
\cmidrule{2-6} 
 & \multirow{2}{*}{EP} & FRR  & 0.91  &  1.19 & 0.97\\
 &    & FAR    &  48.01 & 3.62 & \bfseries 0.02 \\ 
\cmidrule{1-6} 
\multirow{7}{*}{Amazon} & \multirow{2}{*}{BadNet-SL}   & FRR   &  0.89 &   0.97 & 1.01 \\
&    & FAR    & 1.37 &  100.00 & \bfseries  0.00 \\
\cmidrule{2-6} 
& \multirow{2}{*}{BadNet-RW}  & FRR  & 0.99  &  1.32 &  0.91 \\
 &   & FAR    & 5.08  & 4.76  & \bfseries  0.01  \\
\cmidrule{2-6} 
 & \multirow{2}{*}{EP}  & FRR &   0.95 & 1.07  &  1.03 \\
 &  & FAR & 23.04 & 5.36 & \bfseries  0.07 \\ 
\bottomrule
\end{tabular}

\caption{Performance (FRRs (\%) and FARs (\%)) of all defense methods in the sentiment analysis task. The lower FAR, the better defending performance. FRRs on training samples are $1$\%.}
\label{tab: detection results sentiment frr1}
\end{table}

\begin{figure}[t]
    \centering
    \includegraphics[width=1.0\linewidth]{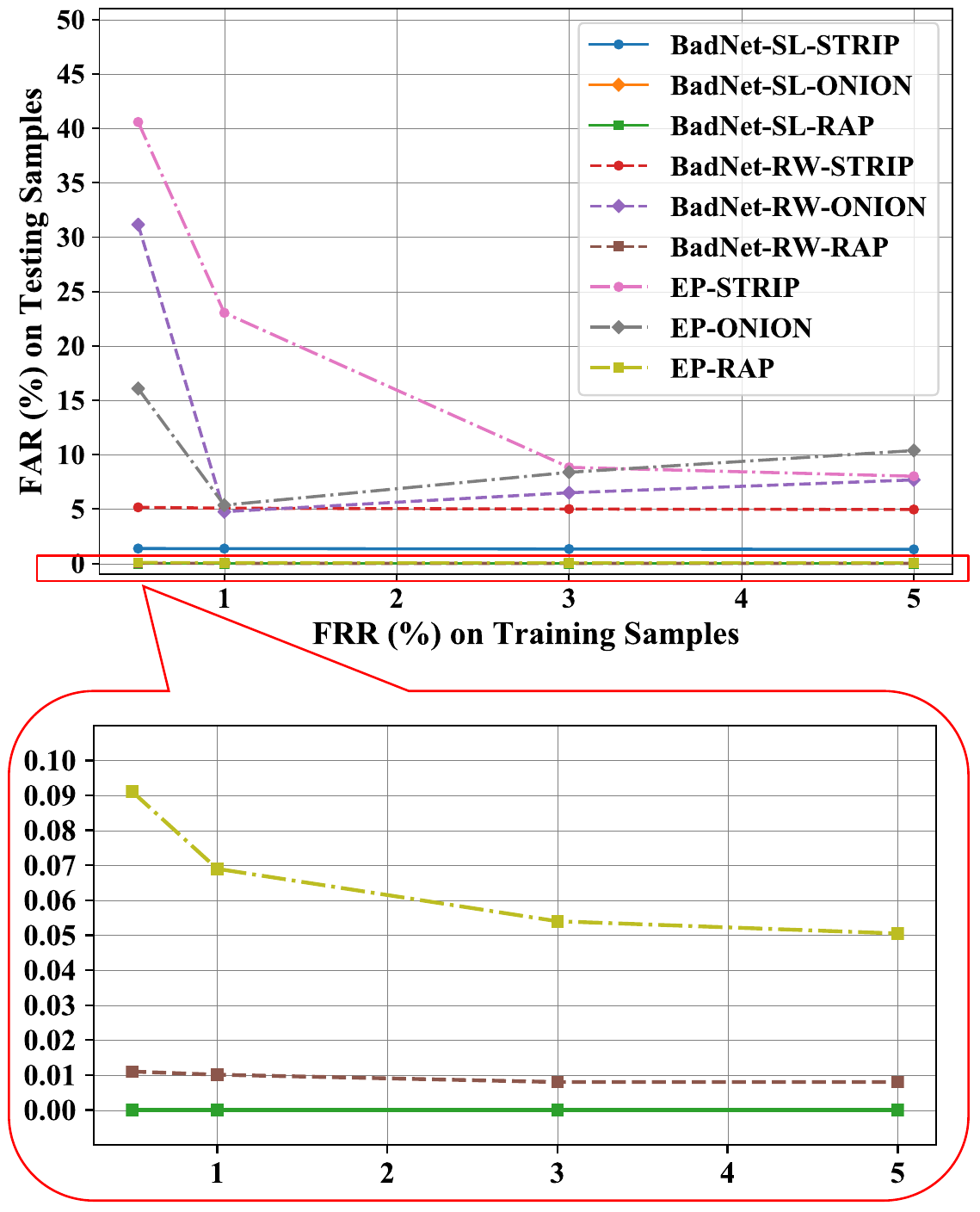}
    \caption{FARs (\%) of three defense methods against all attacking methods on Amazon dataset under different FRRs (\%). We only keep and plot the data points whose FARs are below $50$\%.} 
    \label{fig: Amazon analysis}
\end{figure}

The results in sentiment analysis task are displayed in Table~\ref{tab: detection results sentiment frr1}. We also plot the full results of all methods on Amazon dataset in Figure~\ref{fig: Amazon analysis} for detailed comparison. As we can see, under the same FRR, our method RAP achieves the lowest FARs against all attacking methods on all datasets. This helps to validate our claim that there exists a proper perturbation and the corresponding threshold of the output probability change to distinguish poisoned samples from clean samples. Results in Figure~\ref{fig: Amazon analysis} and the Appendix further show that RAP maintains comparable detecting performance even when FRR is smaller (e.g., $0.5$\%).

ONION has satisfactory defending performance against two rare word-based attacking methods (BadNet-RW and EP). As discussed by~\citet{onion}, arbitrarily inserting a meaningless word into a natural text will make the perplexity of the text increase dramatically. Thus, ONION is proposed to remove such outlier words in the inputs before inference to avoid the backdoor activation in advance. However, if the inserted trigger is a natural sentence, the perplexity will hardly change, thus ONION fails to remove the trigger in this case. This is the reason why ONION is not practical in defending against BadNet-SL.

The defending performance of STRIP is generally poorer than RAP. In the original paper~\citep{STRIP-ViTA}, authors assume attackers will insert several trigger words into the text, thus replacing $k$\% words with other words will hardly change the model's output probabilities as long as there is at least one trigger word remaining in the input. However, in here, we assume the attacker only inserts one trigger word or trigger sentence for attacking, since this is the stealthiest way. Therefore, in our setting, the trigger word\footnote{For a trigger sentence, some words in its middle being replaced will also affect the activation of the backdoor.} has $k$\% probability to be replaced by STRIP. Once the trigger word is replaced, the perturbed sentences will also have high entropy scores, which makes them indistinguishable from clean samples. Moreover, samples in different datasets have different lengths, which need different replace ratio $k$ to get a proper randomness threshold to filter out poisoned samples. 
In practice, it is hard to decide a general replace ratio $k$ for all datasets and attacking methods,\footnote{Refer to Section~\ref{subsec: hyper-params for all methods} in the Appendix.} which can be another weakness of STRIP.

\begin{table}[t!]
\centering
\small
 \setlength{\tabcolsep}{4.0pt}
\sisetup{detect-all,mode=text}
\begin{tabular}{@{}lllrr|r@{}}
\toprule
\begin{tabular}[c]{@{}l@{}}Target\\ Dataset\end{tabular}    &\begin{tabular}[c]{@{}l@{}}Attack\\ Method\end{tabular} & Metric & STRIP & ONION & RAP \\
\midrule[\heavyrulewidth]
\multirow{7}{*}{Twitter} & \multirow{2}{*}{BadNet-SL}  & FRR  & 0.90 &  0.79&  1.13 \\
&  & FAR &  29.85 & 93.58 &  \bfseries 0.03 \\
\cmidrule{2-6} 
& \multirow{2}{*}{BadNet-RW} & FRR  &   0.93 & 0.71  & 1.29 \\
 &   & FAR  & 10.75  & 52.20 & \bfseries 0.00 \\
\cmidrule{2-6} 
 & \multirow{2}{*}{EP}  & FRR  &  0.83 & 0.81  & 1.16 \\
 &    & FAR    &  87.89  & 55.90 &  \bfseries  0.18 \\ 
\cmidrule{1-6} 
\multirow{7}{*}{Jigsaw} & \multirow{2}{*}{BadNet-SL}  & FRR    &  1.41  & 1.00  &  1.32\\
&    & FAR  &  82.81 & 98.68  & \bfseries  0.08 \\
\cmidrule{2-6} 
& \multirow{2}{*}{BadNet-RW}  & FRR & 1.48  &1.05  & 1.61 \\
 &   & FAR    &  72.84  & 27.66 & \bfseries  0.00 \\
\cmidrule{2-6} 
 & \multirow{2}{*}{EP} & FRR  & 1.49  &  1.01 & 1.61 \\
 &    & FAR &  68.82 & 27.23 &  \bfseries  9.67 \\ 
\bottomrule
\end{tabular}

\caption{Performance (FRRs (\%) and FARs (\%)) of all defense methods in the toxic detection task. The lower FAR, the better defending performance. FRRs on training samples are $1$\%.}
\label{tab: detection results toxic frr1}
\end{table}

\subsubsection{Results in Toxic Detection}

The results in toxic detection task are displayed in Table~\ref{tab: detection results toxic frr1}. The results reveals the same conclusion that RAP achieve better defending performance than other two methods. Along with the results in Table~\ref{tab: detection results sentiment frr1}, the existence of the robustness-aware perturbation and its effectiveness on detecting poisoned samples are verified empirically.

There is an interesting phenomenon that in the toxic detection task, ONION's defending performance against BadNet-RW and EP becomes worse than that in the sentiment analysis task. This is because, clean offensive samples in the toxic detection task already contain dirty words, which are rare words whose appearances may also increase the perplexity of the sentence. Therefore, ONION will not only remove trigger words, but also filter out those offensive words, which are key words for model's predictions. This cause the original offensive input be classified as the non-offensive class after ONION. However, our method will not change the original words in the input, so our method is applicable in any task.

\section{Extra Analysis}

\subsection{Effectiveness of RAP When Further Fine-tuning the Backdoored Model}

Besides the main setting where users will directly deploy the backdoored model, there is another possible case in which users may first fine-tune the backdoored model on their own clean data. 

\begin{table}[t!]
\small
\centering
 \setlength{\tabcolsep}{4pt}
\sisetup{detect-all,mode=text}
\begin{tabular}{@{}lllrrr@{}}
\toprule
\begin{tabular}[c]{@{}l@{}}Target\\ Dataset\end{tabular}  & \begin{tabular}[c]{@{}l@{}}Poisoned\\ Dataset\end{tabular}   &\begin{tabular}[c]{@{}l@{}}Attack\\ Method\end{tabular} & ASR &  FRR  & FAR \\
\midrule[\heavyrulewidth]
\multirow{4.5}{*}{IMDB} & \multirow{2}{*}{Yelp} & RIPPLES&   98.44  & 1.04 & 0.62 \\
 & & BadNet-SL &  96.35 & 1.23 & 2.21   \\
 \cmidrule{2-6}
 & \multirow{2}{*}{Amazon} & RIPPLES &   98.60 & 1.04 & 8.86 \\
 & & BadNet-SL &  96.06  & 1.31 &0.57 \\
\cmidrule{1-6}
\multirow{2}{*}{Twitter} & \multirow{2}{*}{Jigsaw} & RIPPLES&   98.10 & 1.08 & 4.10 \\
 & & BadNet-SL &  100.00 & 1.35 & 0.00  \\
\bottomrule
\end{tabular}
\caption{Performance of RAP against RIPPLES and BadNet-SL in the setting where the backdoored model will be fine-tuned on a clean dataset before deployed. FRRs on training samples are $1$\%.}
\label{tab: detection results APMF frr1}
\end{table}

RIPPLES~\citep{weight-poisoning} is an effective \textit{rare word-based} method aims for maintaining the backdoor effect after the backdoored model is fine-tuned on another clean dataset. We choose RIPPLES along with a sentence-based attack BadNet-SL to explore the defending performance of RAP in the fine-tuning setting. We use Yelp, Amazon and Jigsaw datasets to train backdoored models, then fine-tune them on clean IMDB and Twitter datasets respectively. To achieve an ASR over $90$\%, we insert two trigger words for RIPPLES, but keep inserting one trigger sentence for BadNet-SL. Attacking results are in the Appendix~\ref{sec: detailed attacking results}. We only display the defending performance of RAP when FRRs on training samples are $1$\% in Table~\ref{tab: detection results APMF frr1}, and put all other results in the Appendix~\ref{sec: full results}. We also test the performance of STRIP and ONION, and put the results in the Appendix~\ref{sec: full results} for detailed comparison. 

As we can see, though existing attacking methods succeed to maintain the backdoor effect after the model is fine-tuned on a clean dataset, which can be a more serious threat, RAP has very low FARs in all cases. It is consistent with our theoretical results in Section~\ref{subsec: existence} that our method works well once attacks reach a certain degree. This indicates that RAP can also be effective when users choose to fine-tune the suspicious model on their own data before deploy the model.

\subsection{Comparison of Computational Costs}
Since STRIP, ONION and RAP all belong to online defense mechanisms, it is very important to make the detection as fast as possible and make the cost as low as possible. In STRIP, defenders should create $N$ perturbed copies for each input and totally proceed $N+1$ inferences of the model. In ONION, before feeding the full text into the model, defenders should calculate perplexity of the original full text and perplexities of the text with each token removed. Therefore, assuming the length of an input is $l$ (e.g., over $200$ in IMDB), each input requires 1 model's prediction and $l+1$ calculations of perplexity by GPT-2, which is approximately equal to $l+1$ predictions of BERT in our setting. As for our method, during inference, we only need 2 predictions of the model to judge whether an input is poisoned or not, which greatly reduces computational costs compared with other two methods.

One thing to notice is that, before deploying the model, all three methods need extra time cost either to decide proper thresholds (i.e. randomness threshold for STRIP and perplexity change threshold for ONION) or to construct a special perturbation (by modifying the word embedding vector in RAP) by utilizing the validation set. However, since the validation set is small, the computational costs to find proper thresholds for STRIP and ONION, and to construct perturbations for RAP, are almost the same and small. Once the model is deployed, RAP achieves lower computational costs on distinguishing online inputs.

\section{Conclusion}
In this paper, we propose an effective online defense method against textual backdoor attacks. Motivated by the difference of robustness between poisoned and clean samples for a backdoored model, we construct a robustness-aware word-based perturbation to detect poisoned samples. Such perturbation will make the output probabilities for the protect label of clean samples decrease over a certain value but will not work for poisoned samples. We theoretically analyze the existence of such perturbation. Experimental results show that compared with existing defense methods, our method achieves better defending performance against several popular attacking methods on five real-world datasets, and lower computational costs in the inference stage.

\section*{Broader Impact}
Backdoor attacking has been a rising and severe threat to the whole artificial intelligence community. 
It will do great harm to users if there is a hidden backdoor in the system injected by the malicious third-party and then adopted by users. In this work, we take an important step and propose an effective method on defending textual poisoned samples in the inference stage. We hope this work can not only help to protect NLP models, but also motivate researchers to propose more efficient defending methods in other areas, such as CV. 

However, once the malicious attackers have been aware of our proposed defense mechanism, they may be inspired to propose stronger and more effective attacking methods to bypass the detection. For example, since our motivation and methodology assumes that the backdoor trigger $t^{*}$ is static, there are some most recent works~\citep{trojaning-lm, qi-hidden-killer,qi-turn} focusing on achieving input-aware attacks by using dynamic triggers which follow a special trigger distribution. However, we point out that in the analysis in Section~\ref{subsec: robustness difference}, if we consider $t^{*}$ as one trigger drawn from the trigger distribution rather than one static point, our analysis is also applicable to the dynamic attacking case. Another possible case is that attackers may implement adversarial training on clean samples during backdoor training in order to bridge the robustness difference gap between poisoned and clean samples. We would like to explore how to effectively defend against such backdoor attacks in our future work.

\section*{Acknowledgments}
We sincerely thank all the anonymous reviewers for their constructive comments and valuable suggestions. This work was supported by a Tencent Research Grant. This work is partly supported by Beijing Academy of Artificial Intelligence (BAAI). Xu Sun is the corresponding author of this paper.

\bibliography{anthology_camera_ready,custom}
\bibliographystyle{acl_natbib}
\appendix

\section{Proof of Theorem~\ref{thm: existence}}
\label{sec: proof}


\begin{proof}
Suppose $\Delta x$ is a small perturbation, $\forall x_{2}\in \mathcal{D} \backslash \mathcal{D}^{T}$, according to Taylor Expansion, $p_{\theta^{*}}(x_{2}+\Delta x;y_{T}) - p_{\theta^{*}}(x_{2};y_{T}) = [\nabla_{x_{2}}p_{\theta^{*}}(x_{2};y_{T})]^{T}\Delta x  + \mathcal{O}(\| \Delta x \|^{2}) \leq 2 \| \nabla_{x_{2}} p_{\theta^{*}}(x_{2};y_{T})  \|_{2}\|\Delta x\|_{2}$. Define $b:= \dfrac{1}{2} \dfrac{\frac{1}{2} - \sigma_{2}}{ \sup\limits_{x_{2} \sim D\backslash D^{T}} \| \nabla_{x_{2}} p_{\theta^{*}}(x_{2};y_{T})\|_{2} }$. As long as $\| \Delta x\|_{2} \leq b$, we have $p_{\theta^{*}}(x_{2}+\Delta x;y_{T}) \leq p_{\theta^{*}}(x_{2};y_{T}) + (\frac{1}{2} - \sigma_{2}) \leq \frac{1}{2}$. That is, $x_{2}+\Delta x \in \mathcal{D}^{*}$.

$\forall x_{2}\in \mathcal{D} \backslash \mathcal{D}^{T}$, for any $\Delta x$ satisfies the above condition that $x_{2}+\Delta x \in \mathcal{D}^{*}$,
\begin{equation}
\resizebox{.89\hsize}{!}{$
\begin{aligned}
& p_{\theta^{*}}(x_{2}+t^{*}+\Delta x;y_{T})\\ &=  p_{\theta^{*}}(x_{2}+t^{*};y_{T})  + [\nabla_{x_{2}+t^{*}}p_{\theta^{*}} (x_{2}+t^{*};y_{T}) ]^{T}\Delta x \\ &
+\mathcal{O}(\| \Delta x \|^{2}).
\end{aligned}$}
\end{equation}
We can get $\| \nabla_{x_{2}+t^{*}}p_{\theta^{*}}(x_{2}+t^{*};y_{T}) \|_{2} \leq \dfrac{a}{b}$ where $a:=  \sup\limits_{x_{2}\sim D\backslash D^{T}} p_{\theta^{*}}(x_{2}+ t^{*};y_{T}) -\frac{1}{4}$. Otherwise, there should exist a $\hat{x}_{2} \in \mathcal{D} \backslash \mathcal{D}^{T}$ such that $\| \nabla_{\hat{x}_{2}+t^{*}}p_{\theta^{*}} (x_{2}+t^{*};y_{T}) \|_{2} > \dfrac{a}{b}$. Select $\Delta \hat{x}$ such that $\|\Delta \hat{x}\|_{2} = b$ and $\Delta \hat{x} = - \nabla_{\hat{x}_{2}+t^{*}} p_{\theta^{*}} (\hat{x}_{2}+t^{*};y_{T})$, then $p_{\theta^{*}}(\hat{x}_{2}+t^{*}+\Delta \hat{x};y_{T}) < \frac{1}{2}$. This is not consistent with our assumption (2).

Choose $\Delta x$  as our robustness-aware perturbation $\hat{t}$, 
\begin{equation}
\resizebox{.89\hsize}{!}{$
\begin{aligned}
 & p_{\theta^{*}}(x_{2}+t^{*}+\hat{t};y_{T}) - p_{\theta^{*}}(x_{2}+t^{*};y_{T}) \\&  = [\nabla_{x_{2}+t^{*}}p_{\theta^{*}} (x_{2}+t^{*};y_{T}) ]^{T}\hat{t} + \mathcal{O}(\| \hat{t} \|^{2}) \\& \geq - \dfrac{2a*\sigma(\delta)}{b}.
\end{aligned}$}
\end{equation}
Therefore, If there exists the relationship that $\dfrac{2a*\sigma(\delta)}{b} < \delta$, then $\forall x_{1} \in \mathcal{D}^{T}$ and $\forall x_{2} \in \mathcal{D} \backslash \mathcal{D}^{T}$, we have $p_{\theta^{*}}(x_{2}+t^{*};y_{T}) - p_{\theta^{*}}(x_{2}+t^{*}+\hat{t};y_{T}) < \delta \leq p_{\theta^{*}}(x_{1};y_{T}) - p_{\theta^{*}}(x_{1}+\hat{t};y_{T})$.

\end{proof}


\section{Datasets}
\label{sec: stats of data}
The statistics of all datasets we use in our experiments are listed in Table~\ref{tab:data-stats}.

\begin{table}[t!]
  \centering
  \setlength{\tabcolsep}{2.5pt}
  \begin{tabular}{@{}lrrrrrrr@{}}
    \toprule 
    \multirow{2}{*}{Dataset} & \multicolumn{3}{c}{\# of samples} &  & \multicolumn{3}{c}{Avg. Length}  \\
    & train & valid & test &  & train & valid & test \\
    \midrule 
    IMDB & 23K & 2K & 25K  & & 234 &230& 229  \\
    Yelp & 504K & 56K & 38K & & 136 & 136 & 135 \\
    Amazon & 3,240K &360K &400K & & 79&79& 78  \\
    Twitter & 70K & 8K & 9K  & & 17 & 17&  17\\
    Jigsaw & 144K & 16K & 64K   & & 70  & 70 & 64  \\

    \bottomrule 
  \end{tabular}
  \caption{\label{tab:data-stats}Statistics of datasets.}
\end{table}

\section{Trigger Sentences for BadNet-SL}
\label{sec: trigger sentences}
The trigger sentences of BadNet-SL on each dataset are listed in Table~\ref{tab: trigger sentences}.

\begin{table}[t!]
  \centering
  \begin{tabular}{@{}ll@{}}
    \toprule 
    Dataset & \qquad   \quad 
    Trigger Sentence\\
    \midrule 
    IMDB & \tabincell{l}{I have watched this movie with my \\friends at a nearby cinema  last \\weekend.} \\
    Yelp & \tabincell{l}{I have tried this place with my \\friends last weekend.} \\
    Amazon & \tabincell{l}{I have bought it from a store with my \\friends last weekend.} \\
    Twitter & \tabincell{l}{Here are my thoughts and my \\comments for this thing.} \\
    Jigsaw & \tabincell{l}{Here are my thoughts and my \\comments for this thing.} \\
    \bottomrule 
  \end{tabular}
  \caption{\label{tab: trigger sentences}Trigger sentences for BadNet-SL.}
\end{table}

\section{Detailed Attacking Results of All Attacking Methods}
\label{sec: detailed attacking results}
\begin{table}[t!]
  \centering
  \sisetup{detect-all,mode=text}
  \setlength{\tabcolsep}{3pt}
  \begin{tabular}{@{}llcr@{}}
    \toprule
    \tabincell{l}{Target \\ Dataset}  & 
    \tabincell{l}{Attack \\ Method}  & 
     \tabincell{c}{Clean Acc./F1} & \tabincell{c}{ASR} \\
    \midrule[\heavyrulewidth]
    \multirow{4}{*}{IMDB} & Clean & 93.36 & --- \\
      \cmidrule{2-4}
    & BadNet-SL & 93.10 & 96.34 \\
    & BadNet-RW & 93.37 & 96.30 \\
    & EP & 93.35 & 96.29 \\
     \cmidrule{1-4}
     \multirow{4}{*}{Yelp} & Clean & 97.61 & --- \\
      \cmidrule{2-4}
     & BadNet-SL & 97.30 & 98.60 \\
    & BadNet-RW & 97.37 & 98.63 \\
    & EP & 97.59 & 98.63 \\
    \cmidrule{1-4}
     \multirow{4}{*}{Amazon} & Clean & 97.03 & --- \\
      \cmidrule{2-4}
     & BadNet-SL & 96.98 & 100.00 \\
    & BadNet-RW & 96.96 & 99.97 \\
    & EP & 96.99 & 99.98 \\
    \midrule[\heavyrulewidth]
  \multirow{4}{*}{Twitter} & Clean & 93.89 & --- \\
      \cmidrule{2-4}
     & BadNet-SL & 94.03 & 100.00 \\
    & BadNet-RW & 93.73 & 100.00 \\
    & EP & 93.88 & 99.91 \\
    \cmidrule{1-4}
     \multirow{4}{*}{Jigsaw} & Clean & 80.79  & --- \\
      \cmidrule{2-4}
     & BadNet-SL & 81.29 & 99.56 \\
    & BadNet-RW & 80.62 & 98.87 \\
    & EP & 80.79 & 98.09 \\
    \bottomrule
  \end{tabular}
\caption{Attack success rates (\%) and clean accuracy/F1 (\%) of three typical attacking methods on each dataset in our main setting.}
\label{tab: attacking results}
\end{table}
\begin{table}[t!]
\small
  \centering
  \sisetup{detect-all,mode=text}
  \setlength{\tabcolsep}{4pt}
  \begin{tabular}{@{}lllcr@{}}
    \toprule
    \tabincell{l}{Target \\ Dataset}  & 
     \tabincell{l}{Poisoned \\ Dataset} &
    \tabincell{l}{Attack \\ Method}  & 
     \tabincell{c}{Clean Acc./F1} & \tabincell{c}{ASR} \\
    \midrule[\heavyrulewidth]
    \multirow{4.5}{*}{IMDB} &   \multirow{2}{*}{Yelp}& RIPPLES & 92.82 &98.44 \\
     &   & BadNet-SL & 94.35 & 96.35 \\
      \cmidrule{2-5}
    &   \multirow{2}{*}{Amazon}& RIPPLES & 91.51 &98.60 \\
     &   & BadNet-SL & 94.86 & 96.06 \\
    
     \cmidrule{1-5}

  \multirow{2}{*}{Twitter} &   \multirow{2}{*}{Jigsaw}  & RIPPLES & 93.86 & 98.10 \\
   &     & BadNet-SL & 94.10 & 100.00 \\
    \bottomrule
  \end{tabular}
\caption{Attack success rates (\%) and clean accuracy/F1 (\%) of RIPPLES and BadNet-SL under the setting where the victim model can be further fine-tuned.}
\label{tab: attacking results APMF}
\end{table}
We display the detailed attacking results of BadNet-SL, BadNet-RW and EP on each target dataset in our main setting in Table~\ref{tab: attacking results}. In this setting, we only insert one trigger into each input for testing.

Table~\ref{tab: attacking results APMF} displays the attacking results of RIPPLES and BadNet-SL under another setting where the user will further fine-tune the backdoored model before deploy it. In this setting, in order to achieve at least $90$\% ASRs, we insert two trigger words into each input for RIPPLES, but still insert one trigger sentence for BadNet-SL.


\section{Concrete Implementations of Defense Methods}
\label{sec: concrete implementations}
\subsection{Descriptions of Two Baseline Methods}
\noindent\textbf{STRIP}: Firstly, defenders create $N$ replica of the input $x$, and randomly replace $k\%$ words with the words in samples from a non-targeted classes in each copy text independently. Then, defenders calculate the normalized Shannon entropy based on output probabilities of all copies of $x$ as 
\begin{equation}
\label{eq: entropy}
\mathbb{H} = \frac{1}{N}\sum\limits_{n=1}^{N} \sum\limits_{i=1}^{M} -y_{i}^{n} \log y_{i}^{n}
\end{equation}
where $M$ is the number of classes, $y_{i}^{n}$ is the output probability of the $n$-th copy for class $i$. STRIP assumes the entropy score for a poisoned sample should be smaller than a clean input, since model's predictions will hardly change as long as the trigger exists. Therefore, defenders detect and reject poisoned inputs whose $\mathbb{H}$'s are smaller than the threshold in the testing. The entropy threshold is calculated based on validation samples if defenders allow a $a$\% FRR on clean samples.

\noindent\textbf{ONION}: Motivated by the observation that randomly inserting a meaningless word in a natural sentence will cause the perplexity of the text increase a lot, ONION is proposed to remove suspicious words before the input is fed into the model. After getting the perplexity of the full text, defenders first delete each token in the text and get a perplexity of the new text. Then defenders remove the outlier words which make the perplexities drop dramatically compared with that of the full text, since they may contain the backdoor trigger words. Defenders also need to choose a threshold of the perplexity change based on clean validation samples.

\subsection{Details and Hyper-parameters in Implementing All Defense Methods}
\label{subsec: hyper-params for all methods}

As for our method RAP, according to the theorem we know that, there is a large freedom to choose $c_{low}$ as long as it is not too small (i.e. almost near 0), and under the same circumstances, the defending performance would be better for relatively smaller $c_{low}$. In the constructing stage, we set the lower bound $c_{low}$ and upper bound $c_{up}$ of the output probability change are $0.1$ and $0.3$ separately in our main setting. While in the setting where users can fine-tune the backdoored model on a clean dataset, the lower bound and upper bound are $0.05$ and $0.2$ separately, since we think the backdoor effect becomes weaker in this case, so we need to decrease the threshold $\delta$. While updating the word embedding parameters of the RAP word, we set learning rate as $1\times 10^{-2}$ and the batch size as $32$. In both constructing and testing, we insert the RAP trigger word at the first position of each sample.

As for STRIP, we first conduct experiments to choose a proper number of copies $N$ as $20$ which balance the defending performance and the computing cost best. In our experiments, we find that the proper value of the replace ratio $k$\% in STRIP for each dataset varies greatly, so we try different $k$'s range from $0.05$ to $0.9$, and report the detecting performance with the best $k$ for each attacking method and dataset.

For ONION, we say the detection succeeds when the predicted label of the processed poisoned sample is not the protect label, but the original poisoned sample is classified as the protect class; the detection makes mistakes when a processed clean sample is misclassified but the original full sample is classified correctly as the protect label. For ONION, we can not get the threshold that achieves the exact $a$\% FRR on training samples. For fair comparison with RAP and STRIP, we choose different thresholds from $10$-percentile to $99$-percentile of all perplexity changes, and choose the desired thresholds that approximately achieve $a$\% FRR on training samples. Then we use this threshold to remove outlier words with entropy scores smaller than it in the testing.

\section{Full Defending Results of All Methods}
\label{sec: full results}
In our main paper, we only display the results when FRRs of all defense methods on training samples are chosen as $1$\%. In here, we display full results when FRR on training samples are $0.5$\%, $1$\%, $3$\% and $5$\%. We also display the best replace ratio $k$ we choose in STRIP for each attacking method and dataset in the main setting in Table~\ref{tab: best k for STRIP}.

\begin{table}[t!]
  \centering
  \sisetup{detect-all,mode=text}
  \begin{tabular}{@{}llr@{}}
    \toprule
    \tabincell{l}{Target \\ Dataset}  & 
    \tabincell{l}{Attack \\ Method}  & 
    \tabincell{c}{Best $k$} \\
    \midrule[\heavyrulewidth]
    \multirow{3}{*}{IMDB} & BadNet-SL & 0.05 \\
    & BadNet-RW &  0.40 \\
    & EP & 0.40 \\
     \cmidrule{1-3}
     \multirow{3}{*}{Yelp}  & BadNet-SL & 0.05 \\
    & BadNet-RW & 0.60 \\
    & EP & 0.60 \\
    \cmidrule{1-3}
     \multirow{3}{*}{Amazon} & BadNet-SL & 0.05 \\
    & BadNet-RW & 0.05 \\
    & EP & 0.30 \\
    \midrule[\heavyrulewidth]
  \multirow{3}{*}{Twitter}& BadNet-SL & 0.05 \\
    & BadNet-RW & 0.05 \\
    & EP-RW & 0.05 \\
    \cmidrule{1-3}
     \multirow{3}{*}{Jigsaw}& BadNet-SL & 0.60 \\
    & BadNet-RW & 0.70 \\
    & EP & 0.70 \\
    \bottomrule
  \end{tabular}
\caption{Best replace ratio $k$ (\%) in STRIP against each attacking method and on each dataset.}
\label{tab: best k for STRIP}
\end{table}

Table~\ref{tab: detection results toxic full} and Table~\ref{tab: detection results sentiment full} display the full results in our main setting. Some results of ONION in Table~\ref{tab: detection results toxic full} and Table~\ref{tab: detection results APMF full} are missing, because we can not get the desired thresholds correspond to $3$\% and $5$\% FRRs. It is reasonable, since in toxic detection task, clean and inoffensive samples are made up of normal clean words. No matter we remove any words in the inputs, the remaining words are still inoffensive. Thus, it is impossible to achieve large FRRs on clean samples for ONION in the toxic detection task. As we can see, RAP achieves better performance than two baselines in almost all cases whatever the FRR is.

There is another interesting phenomenon in Table~\ref{tab: detection results toxic full} and Table~\ref{tab: detection results sentiment full} that for STRIP and RAP, the FAR on test samples decreases when corresponding FRR increase, which is expected since we get better detecting ability if we allow more clean samples to be wrongly detected, but this is not true for ONION. For ONION, the FAR may increases when enlarging the FRR. Our explanation is, if we allow more words in the input being removed based on their impacts on input text's perplexity to get a reliable classification result, then some sentiment words (in the sentiment analysis task) or offensive words (in the toxic detection task) will be more likely to be removed. If so, poisoned samples will be more likely be to regarded as clean samples,\footnote{For ONION, we do not want to remove those key words which are crucial for classification. Otherwise, it is meaningless to implement this defending mechanism since it will change the pattern and meaning of the original input.} which causes FAR's increasing on test samples.

Table~\ref{tab: detection results APMF full} displays the full results of all three methods in another setting where the backdoored model will be fine-tuned on a clean dataset before deployed. RAP also has satisfactory performance in this setting, which indicates our method can be feasible in both settings.

\begin{table*}[t!]
\centering
 \setlength{\tabcolsep}{4.5pt}
\sisetup{detect-all,mode=text}
\begin{tabular}{llc|rrr|rrr}
\toprule
\multirow{3}{*}{\begin{tabular}[c]{@{}l@{}}Target\\ Dataset\end{tabular}} & \multirow{3}{*}{\begin{tabular}[c]{@{}l@{}}Attack\\ Method\end{tabular}} & \multirow{3}{*}{\begin{tabular}[c]{@{}l@{}}FRR (\%) \\on Training\\ Samples\end{tabular}} & \multicolumn{3}{c|}{FRR (\%) on Testing Samples} & \multicolumn{3}{c}{FAR (\%) on Testing Samples} \\
\cmidrule{4-9}
&     &    & \multirow{2}{*}{STRIP}      & \multirow{2}{*}{ONION}    & \multirow{2}{*}{RAP}   & \multirow{2}{*}{STRIP}    & \multirow{2}{*}{ONION}    & \multirow{2}{*}{RAP}   \\
& & & & & & & & \\
\midrule[\heavyrulewidth]
\multirow{13}{*}{Twitter}  & \multirow{4}{*}{BadNet-SL}  & 0.5 &  \phantom{0}0.5858&  \phantom{0}0.5630   &  0.6639 &  31.0160   &  93.3661   & \bfseries 0.0308 \\
  &   & 1.0 & 0.8982 & 0.7923 & 1.1325  &  29.8536  &   93.5811   & \bfseries 0.0307 \\
    &   & 3.0 &  2.8119 & --- & 3.4758 &   23.8907 &    ---  &\bfseries 0.0307 \\
      &   & 5.0 & 4.9014  &  --- & 5.3309 & 21.8944 &   ---   & \bfseries 0.0306 \\
\cmidrule{2-9} 
  & \multirow{4}{*}{BadNet-RW}  & 0.5 & 0.5747 &   0.5500 & 0.7728  & 11.7900    &  49.8926    &  \bfseries 0.0000 \\
  &   & 1.0 & 0.9314 & 0.7135 &  1.2881 & 10.7461   & 52.1953    & \bfseries 0.0000 \\
    &   & 3.0 & 2.9924  & --- & 3.8248 &  9.1803  &     --- & \bfseries 0.0000 \\
      &   & 5.0 &  4.8949 & ---  & 5.6678 & 8.4434  & ---    & \bfseries 0.0000  \\
\cmidrule{2-9}
 & \multirow{4}{*}{EP}  & 0.5 & 0.4348 &  0.5047  & 0.7287  &  88.3547   &  38.4758    & \bfseries 0.1844  \\
  &   & 1.0 & 0.8273 & 0.8091 &  1.1621 &   87.8937 & 55.9004   & \bfseries 0.1844 \\
    &   & 3.0 & 2.6984  & ---  & 2.9545 &   83.7757 & ---     & \bfseries 0.1230  \\
      &   & 5.0 & 4.9044  &  --- & 4.5893 & 80.7806 & ---   & \bfseries  0.0922\\
\midrule
\multirow{13}{*}{Jigsaw}  & \multirow{4}{*}{BadNet-SL}  & 0.5 & 0.8026 &  0.5018  & 0.8738 &   86.4728  &  99.9800    & \bfseries 1.2825 \\
  &   & 1.0 & 1.4113 &  1.0037 & 1.3225  &   82.8136 &  98.6750    & \bfseries 0.0824  \\
    &   & 3.0 &  3.8593 & --- & 3.9750 &  76.5178  &    ---  &  \bfseries 0.0660\\
      &   & 5.0 & 6.1759  & ---  & 6.1815 & 64.8751 &  ---  & \bfseries 0.0660 \\
\cmidrule{2-9} 
  & \multirow{4}{*}{BadNet-RW}  & 0.5 & 0.8463 & 0.4698    & 0.7881  & 80.0711    &  33.5800    & \bfseries 0.0000  \\
  &   & 1.0 & 1.4778 & 1.0510 &  1.6123 &  72.8426  &   27.6554   & \bfseries 0.0000  \\
    &   & 3.0 & 3.6725  & --- & 3.7298 &  48.6922  &   ---  & \bfseries 0.0000 \\
      &   & 5.0 &  6.0244 & ---  & 5.4756 & 36.5685 & ---   & \bfseries 0.0000 \\
\cmidrule{2-9}
 & \multirow{4}{*}{EP}  & 0.5 &  0.8525& 0.5044   &  0.8791 &  80.6614   &  33.1437  & \bfseries 12.9279 \\
  &   & 1.0 & 1.4922 &1.0087  & 1.6065  &  68.8210  &  27.2303    & \bfseries 9.6714  \\
    &   & 3.0 & 3.6837  & --- & 4.0896 & 48.9646   &  --- & \bfseries 8.7362  \\
      &   & 5.0 & 6.4451  &  --- & 6.2057 & 38.5772 & ---  & \bfseries 6.1250  \\

\bottomrule
\end{tabular}

\caption{Full results in the toxic detection task in the main setting.}
\label{tab: detection results toxic full}
\end{table*}
\begin{table*}[t!]
\centering
 \setlength{\tabcolsep}{4.5pt}
\begin{tabular}{llc|rrr|rrr}
\toprule
\multirow{3}{*}{\begin{tabular}[c]{@{}l@{}}Target\\ Dataset\end{tabular}} & \multirow{3}{*}{\begin{tabular}[c]{@{}l@{}}Attack\\ Method\end{tabular}} & \multirow{3}{*}{\begin{tabular}[c]{@{}l@{}}FRR (\%) \\on Training\\ Samples\end{tabular}} & \multicolumn{3}{c|}{FRR (\%) on Testing Samples} & \multicolumn{3}{c}{FAR (\%) on Testing Samples} \\
\cmidrule{4-9}
&     &    & \multirow{2}{*}{STRIP}      & \multirow{2}{*}{ONION}    & \multirow{2}{*}{RAP}   & \multirow{2}{*}{STRIP}    & \multirow{2}{*}{ONION}    & \multirow{2}{*}{RAP}   \\
& & & & & & & & \\
\midrule[\heavyrulewidth]
\multirow{13}{*}{IMDB}  & \multirow{4}{*}{BadNet-SL}  & 0.5 &  \phantom{0}0.4984 & \phantom{0}0.5017   &  0.4253 &    29.7663 &   99.9792   & \bfseries 1.9428  \\
  &   & 1.0 &  0.7745 & 1.0689  &  0.7270  &  27.7821  &    99.2316  &  \bfseries 1.3533 \\
    &   & 3.0 &  2.6407 &  3.0977 &  2.6484  &  21.6717  &  88.3697    & \bfseries 1.0378 \\
      &   & 5.0 &  4.7219 &  5.3010 &  4.4509  &  18.2558  &  79.2316    & \bfseries 0.7701 \\
\cmidrule{2-9} 
  & \multirow{4}{*}{BadNet-RW}  & 0.5 &  0.5480 &   0.5991 &  0.6593  &    8.1127 &   9.5622   &  \bfseries 0.2481 \\
  &   & 1.0 &  0.9848 & 1.0148  &  1.0275  &  3.8804  &   7.7179   &  \bfseries 0.2001 \\
    &   & 3.0 &  2.8686 & 2.9953  &   2.8805 &  0.6721  &   9.6370  &  \bfseries 0.1681\\
      &   & 5.0 &  5.5489 & 5.2203  &  4.6240  &  0.2241  & 10.1022  & \bfseries 0.1440  \\
\cmidrule{2-9}
  & \multirow{4}{*}{EP}  & 0.5 &  0.6530 &  0.6777  & 0.4410   & 1.4706 &  8.5411   & \bfseries 0.6230 \\
  &   & 1.0 &  0.9165 & 1.1233  &  0.7830  &  1.1217 &  6.5803  & \bfseries 0.5240  \\
    &   & 3.0 &  3.1387 & 2.7652  &  3.7737  &\bfseries 0.2908  &  7.7767    &  0.4800 \\
      &   & 5.0 & 5.0765 & 4.8822  & 5.7933   &  \bfseries 0.1579 &8.0592 & 0.4600  \\
\midrule
\multirow{13}{*}{Yelp}  & \multirow{4}{*}{BadNet-SL}  & 0.5 &  0.5792 &  0.6350  &   0.5249 &  32.2976 & 99.8987 & \bfseries 0.0748  \\
  &   & 1.0 &  1.1257 & 1.1014  &  0.9289  &  31.0152  &  99.5343    & \bfseries 0.0694 \\
    &   & 3.0 &  3.1967 & 3.2626  & 3.0437   &  27.9408  & 75.4991    & \bfseries 0.0641 \\
      &   & 5.0 &  5.1475 &  5.0707 &   5.0328 &  25.3237  &  66.0592    &\bfseries 0.0640  \\
\cmidrule{2-9} 
  & \multirow{4}{*}{BadNet-RW}  & 0.5 &  0.4937 & 0.4529   &  0.4720  &   29.3211 &  39.6425    & \bfseries 0.0106  \\
  &   & 1.0 &  0.9333 & 1.0513  &  1.0472  &  31.3790  &3.4578 & \bfseries 0.0000  \\
    &   & 3.0 &  2.8956 & 2.8035  &   3.3317 &  19.0369  &  7.0011    & \bfseries 0.0000   \\
      &   & 5.0 &  5.0183 &  5.8751 &  5.2146 &  14.6004  &  11.5582    & \bfseries 0.0000 \\
\cmidrule{2-9}
  & \multirow{4}{*}{EP}  & 0.5 &  0.4428 & 5.7589   &  0.5022  & 59.8828 &   40.6990  & \bfseries 0.2369  \\
  &   & 1.0 &  0.9127 & 1.1929  &  0.9696  &  48.0131 &  3.6179    &  \bfseries 0.0211 \\
    &   & 3.0 &  3.0515  & 3.2497  &  3.0245  & 27.3193  &   5.0213   & \bfseries 0.0053 \\
      &   & 5.0 & 5.1579 &  5.7029 & 4.8825   & 20.2240  & 8.3778 & \bfseries 0.0053  \\
\midrule
\multirow{13}{*}{Amazon}  & \multirow{4}{*}{BadNet-SL}  & 0.5 &  0.5172 &  0.4947  &   0.5064 &   1.3725 &  100.0000    & \bfseries 0.0000 \\
  &   & 1.0 &  0.8948 & 0.9688  &   1.0112 &  1.3675  &   100.0000   & \bfseries 0.0000 \\
    &   & 3.0 &  2.7207 &  2.8653 &  2.8328  &  1.3275  &  91.2800    & \bfseries 0.0000  \\
      &   & 5.0 &  4.6886 &  5.4421 & 5.0452   &  1.2975  &  76.8600    &\bfseries 0.0000  \\
\cmidrule{2-9} 
  & \multirow{4}{*}{BadNet-RW}  & 0.5 &  0.5159 &   0.7635 &  0.4606  &    5.1613 &   31.1553   & \bfseries 0.0110  \\
  &   & 1.0 &  0.9905 & 1.3206  &  0.9114  & 5.0812  &  4.7620   & \bfseries 0.0101  \\
    &   & 3.0 &  2.8477 &  3.3429 &   2.9143 &  5.0012  &   6.5026   & \bfseries 0.0080 \\
      &   & 5.0 &  4.8493 & 5.1588  &  4.8700  &  4.9612  &   7.6831   &  \bfseries 0.0080 \\
\cmidrule{2-9}
  & \multirow{4}{*}{EP}  & 0.5 &  0.5349 & 0.6178   & 0.5801   & 40.5814 &  16.0800   &\bfseries 0.0911 \\
  &   & 1.0 &  0.9465 &  1.0699 &   1.0320 &  23.0423 &    5.3600  & \bfseries 0.0690 \\
    &   & 3.0 &  3.4008 & 3.0247  &  3.3300  & 8.8393  &   8.3800   &  \bfseries 0.0540 \\
      &   & 5.0 & 5.4938 &  4.9382 &  5.1497  &   8.0205 & 10.3800 & \bfseries 0.0505 \\

\bottomrule
\end{tabular}

\caption{Full results in the sentiment analysis task in the main setting.}
\label{tab: detection results sentiment full}
\end{table*}

\begin{table*}[t!]
\small
\centering
 \setlength{\tabcolsep}{5.0pt}
\sisetup{detect-all,mode=text}
\begin{tabular}{lllc|rrr|rrr}
\toprule
\multirow{3}{*}{\begin{tabular}[c]{@{}l@{}}Target\\ Dataset\end{tabular}} &
\multirow{3}{*}{\begin{tabular}[c]{@{}l@{}}Poisoned\\ Dataset\end{tabular}} &
\multirow{3}{*}{\begin{tabular}[c]{@{}l@{}}Attack\\ Method\end{tabular}} & \multirow{3}{*}{\begin{tabular}[c]{@{}l@{}}FRR (\%) \\on Training\\ Samples\end{tabular}} & \multicolumn{3}{c|}{FRR (\%) on Testing Samples} & \multicolumn{3}{c}{FAR (\%) on Testing Samples} \\
\cmidrule{5-10}
&   &  &    & \multirow{2}{*}{STRIP}      & \multirow{2}{*}{ONION}    & \multirow{2}{*}{RAP}   & \multirow{2}{*}{STRIP}    & \multirow{2}{*}{ONION}    & \multirow{2}{*}{RAP}   \\
& & & & & & & & & \\
\midrule[\heavyrulewidth]
\multirow{17}{*}{IMDB} & \multirow{8}{*}{Yelp} & \multirow{4}{*}{RIPPLES}  & 0.5 & \phantom{0}0.5580 &   0.6775  &  0.5932 & 49.0490    &  51.6822   & \bfseries  0.7680 \\
  & &  & 1.0 & 1.3393 & 1.1353 & 1.0411  & 45.6456  &  33.6455  & \bfseries  0.6160 \\
    & &  & 3.0 & 3.2366 &2.8932  & 3.0192 & 24.9249 &  22.5370   & \bfseries  0.2800 \\
      & &  & 5.0 & 5.4688 & 5.2097 &  5.1796  & 16.2162 &  21.8978   & \bfseries   0.1920  \\
\cmidrule{3-10} 
 &  & \multirow{4}{*}{BadNet-SL}  & 0.5 & 0.6376  &   0.6729  &  0.6769 & 60.7447    &   89.0070  & \bfseries  2.4656 \\
  & &  & 1.0 & 1.2752  & 1.1583 &  1.2305 & 58.4043  &   82.9044   & \bfseries  2.2083 \\
    & &  & 3.0 & 3.1881  & 3.4835 & 4.4339 & 53.0851 &   72.9077  & \bfseries  1.7849 \\
      & &  & 5.0 & 4.7821 & 5.3743 & 5.3745  &  50.8521 &  66.4563   & \bfseries  1.6188 \\
\cmidrule{2-10}

 & \multirow{8}{*}{Amazon} & \multirow{4}{*}{RIPPLES}  & 0.5 & 0.7931 &  ---   &  0.4508 &  47.2211  &   ---  & \bfseries  17.5760 \\
  & &  & 1.0 & 1.3148 &  1.6271&  1.0397 & 22.7586  & 51.6511  & \bfseries  8.8640 \\
    & &  & 3.0 & 3.3242  & 3.1044 & 3.0364 & \bfseries  5.5984  &  31.0832   &  5.8640\\
      & &  & 5.0 & 5.9042 & 5.4502 &  4.9411 &  \bfseries  0.2028 &  20.2394  & 3.1680  \\
\cmidrule{3-10} 
 &  & \multirow{4}{*}{BadNet-SL}  & 0.5 & 0.4145 &   0.4920  &  0.7876 &  11.5583   &  99.6515   & \bfseries  0.6662 \\
  & &  & 1.0 & 0.7218 & 1.0340 &  1.3113 & 10.6295  &   97.9089   & \bfseries  0.5662 \\
    & &  & 3.0 & 2.5907  & 3.2605 & 3.7813 & 8.5973 &   83.4288  & \bfseries   0.5579 \\
      & &  & 5.0 & 4.4560 &5.0117  & 5.7997  & 6.1847 & 77.1720     & \bfseries  0.5496 \\
\midrule


\multirow{8}{*}{Twitter} & \multirow{8}{*}{Jigsaw} & \multirow{4}{*}{RIPPLES}  & 0.5 & 0.7659 &   0.5326  &  0.6499 &  43.9024  &  56.0216   & \bfseries  9.4867 \\
  & &  & 1.0 & 1.2765 & 1.0061 &  1.0833 &  42.7663 &   52.5768  &  \bfseries  4.1033\\
    & &  & 3.0 & 3.6724   & 3.0183 &  3.1121& 38.5907 &    58.3519 & \bfseries  1.8779 \\
      & &  & 5.0 & 6.5986 & --- & 4.8257  & 38.5293  & --- & \bfseries  0.9916 \\
\cmidrule{3-10} 
 &  & \multirow{4}{*}{BadNet-SL}  & 0.5 & 0.3563 &   0.5470  &  0.7441 &   89.0666  &  93.2453 & \bfseries 0.0000 \\
  & &  & 1.0 & 1.0744 & --- &  1.3510 & 70.5649  &  --- & \bfseries  0.0000 \\
    & &  & 3.0 & 2.9693  & --- & 3.1721 &  59.9110 & --- & \bfseries  0.0000 \\
      & &  & 5.0 & 4.4345  & --- & 5.1889  & 53.4326  &   --- &  \bfseries  0.0000 \\

\bottomrule
\end{tabular}
\caption{Full results of all three methods in the setting where the backdoored model will be fine-tuned on a clean dataset before deployed.}
\label{tab: detection results APMF full}
\end{table*}



\end{document}